\documentclass{article}

\usepackage{microtype}
\usepackage{graphicx}
\usepackage{subfigure}
\usepackage{booktabs} % for professional tables
\usepackage{dirtytalk}

\usepackage{hyperref}

\usepackage[accepted]{icml2021}

\icmltitlerunning{Sensitivity analysis in differentially private machine learning using hybrid automatic differentiation}

\begin{document}

\twocolumn[
\icmltitle{Sensitivity analysis in differentially private machine learning using hybrid automatic differentiation}



\icmlsetsymbol{equal}{*}

\begin{icmlauthorlist}
\icmlauthor{Alexander Ziller}{tumai,tumrad,om}
\icmlauthor{Dmitrii Usynin}{tumai,tumrad,icl,om}
\icmlauthor{Moritz Knolle}{tumai,tumrad}
\icmlauthor{Kritika Prakash}{om}
\icmlauthor{Andrew Trask}{om}
\icmlauthor{Rickmer Braren}{tumrad}
\icmlauthor{Marcus Makowski}{tumrad}
\icmlauthor{Daniel Rueckert}{tumai,icl,om}
\icmlauthor{Georgios Kaissis}{tumai,tumrad,icl,om}
\end{icmlauthorlist}

\icmlaffiliation{tumai}{Artificial Intelligence in Medicine and Healthcare, Technical University of Munich, Munich, Germany}
\icmlaffiliation{tumrad}{Institute of Diagnostic and Interventional Radiology, Technical University of Munich, Munich, Germany}
\icmlaffiliation{icl}{Department of Computing, Imperial College London, London, United Kingdom}
\icmlaffiliation{om}{OpenMined}

\icmlcorrespondingauthor{Georgios Kaissis}{g.kaissis@tum.de}
% \icmlcorrespondingauthor{Eee Pppp}{ep@eden.co.uk}

\icmlkeywords{Machine Learning, Differential Privacy, ICML}

\vskip 0.3in
]



\printAffiliationsAndNotice{}  % leave blank if no need to mention equal contribution

\begin{abstract}
In recent years, formal methods of privacy protection such as differential privacy (DP), capable of deployment to data-driven tasks such as machine learning (ML), have emerged. Reconciling large-scale ML with the closed-form reasoning required for the principled analysis of individual privacy loss requires the introduction of new tools for automatic sensitivity analysis and for tracking an individual's data and their features through the flow of computation. For this purpose, we introduce a novel \textit{hybrid} automatic differentiation (AD) system which combines the efficiency of reverse-mode AD with an ability to obtain a closed-form expression for any given quantity in the computational graph. This enables modelling the sensitivity of arbitrary differentiable function compositions, such as the training of neural networks on private data. We demonstrate our approach by analysing the individual DP guarantees of statistical database queries. Moreover, we investigate the application of our technique to the training of DP neural networks. Our approach can enable the principled reasoning about privacy loss in the setting of data processing, and further the development of automatic sensitivity analysis and privacy budgeting systems.

\end{abstract}

\section{Introduction}
The handling and processing of sensitive human data inherently entails the risk of compromising individual privacy and exposing unwarranted quantities of personal information. After a widespread enforcement of personal data regulations as well as the emergence of an increasingly privacy-conscious population, the requirement has arisen to augment machine learning (ML) systems with a number of algorithmic solutions for private, regulation-compliant training. Extensions to popular ML libraries and deep learning toolkits enabling DP deep learning provide a solution by retrofitting established DP mechanisms to ML workflows.\newline
The most successful application of DP to deep learning has arguably been the DP-SGD algorithm \cite{abadi2016deep}, which, although empirically effective in many cases, quintessentially relies on imposing a bound on the sensitivity by clipping gradient norms. The resulting geometric bias is associated with a loss of utility of the final model which may exceed the utility penalty of noise addition alone. The very requirement for gradient clipping is a result of a limited ability for introspection of the training process: At any time during training, some quantity in the network (input, label, weight, etc.) may cause an unbounded growth of the gradient norm. Moreover, recent work has highlighted that the mere application of DP mechanisms such as DP-SGD to existing model architectures may not lead to optimal outcomes, as DP deep learning introduces specific requirements which may be served by different architectural choices than learning in the non-private setting \cite{papernot2019making}. Such choices should (a) be made in a principled way and (b) should optimally not require the utilisation of private data. Current deep learning frameworks (and their DP extensions) are designed with efficient computations over batches of input data in mind. Neither the decoupling of model development from the actual data processing nor the ability to easily obtain closed-form mathematical expressions is inherent to their design. \newline
Even beyond deep learning, in the setting of statistical learning and queries over databases, recent work \cite{feldman2020individual} showcases that introspecting the properties of the functions applied to private database records can provide improved privacy guarantees to the individual. \newline
From the above, we identify three key requirements for new computational tools in the realm of private machine learning: (1) An improved ability to model the properties of function compositions over private data and, especially, their influence on sensitivity; (2) a design philosophy centred on the data of a single individual and their features; (3) the option to reason about an algorithm's privacy characteristics without having to input sensitive data. Our work tackles these requirements by introducing a novel \textit{hybrid} automatic differentiation (AD) framework. It combines the efficiency of reverse-mode AD over computational graphs with the ability to obtain a closed-form expression for every quantity present in the graph. It can thus be used to quantify the privacy characteristics of algorithms used to analyse sensitive data. \newline
The rest of the paper is organised as follows:
\begin{itemize}
    \item Section \ref{related_work} provides an overview of key terms used and related work
    \item Section \ref{hybrid_autodiff} gives an overview of our proposed hybrid automatic differentiation system
    \item Section \ref{experiments} exemplifies our system in the context of statistical queries and the DP-SGD algorithm 
    \item We conclude with a concise discussion of limitations and future directions in Section \ref{discussion}
\end{itemize}

\section{Related work}
\label{related_work}

\paragraph{Automatic and symbolic differentiation}
Automatic differentiation (AD) is a method to automatically determine the differentials of expressions with respect to their components to computer precision. In deep neural networks \textit{reverse mode} automatic differentiation is used which is computationally efficient when expressions with multiple inputs have a scalar output. Forward-mode automatic differentiation also exists, but is not covered in this work. On a high level, AD stores computations in a graph data structure in which every node maintains a reference to its parents as well as the operation from which it originated. The partial derivatives of these primitive operations are then utilised by repeated application of the chain rule of calculus over the graph sorted in reverse topological order (\textit{reverse accumulation step}) to obtain the differentials (or gradients) with respect to the leaves of the graph. Modified versions of this system exist, but are concentrated on memory efficiency in the context of computational graph storage. We refer the reader to \cite{baydin2018automatic} for an overview. AD is a key component in deep learning frameworks, where it is utilised \say{behind the scenes} to perform the backpropagation step. Hence, the user only need specify the forward path of computation (from inputs to e.g. the loss calculation), a step often referred to as \say{graph generation}, since the computational graph is defined (either ahead of time or at run-time) by the programmatic inputs of the user.\newline
Symbolic differentiation (SD) differs from automatic differentiation in that it allows the calculation of derivatives in an analytical (closed) form by manipulating mathematical expressions under a constrained set of rules. It is commonly utilised in symbolic algebra systems \cite{sympy}, focused more on algebraic manipulation than on the complex neural network training. Therefore, most existing systems do not provide a simple method to specify such complex architectures. Beyond this design choice, SD suffers from a phenomenon termed \textit{expression swell}, in which derivatives of large functions become computationally impractical to store in memory and very slow to analytically calculate. To our knowledge, no SD system is widely utilised for training neural networks. \newline
The closest line of work to our system are \textit{static-graph-based} systems such as \textit{Aesara}, a successor to the \textit{Theano} differentiable programming framework \cite{bergstra2011theano}. Such systems are typically optimised for tensor operations, as they have been designed to allow efficient gradient computations for deep learning or gradient-assisted \textit{Monte Carlo} sampling. Thus, their focus is not the provision of closed-form mathematical  expressions for functions in the network.
\paragraph{Differential Privacy and DP-SGD}
We assume familiarity with Differential Privacy and hence omit a detailed discussion at this point, however refer to \cite{Dwork2013}. We will use the definition of DP as put forward in this work unless noted otherwise. \newline
The application of DP to the training of deep neural networks was described in \cite{abadi2016deep} and termed DP-SGD. We reproduce some key insights of this work here to motivate our discussion below: \newline
On a high level, the \textit{mechanism} applied to private data in DP-SGD is a deep neural network, whose output, a scalar value termed the \textit{loss}, is differentiated with respect to the weights to obtain an update rule. As deep neural networks do not necessarily satisfy any assumptions about \textit{Lipschitz} continuity of their gradients and the gradient's $L_2$-norm represents the \textit{sensitivity} term, gradients are processed by \textit{clipping}, that is, bounding them to a desired $L_2$-norm, proportional to which Gaussian noise is added to satisfy DP. The clipping step introduces geometric bias to the gradient \cite{chen2020understanding}, which is undesirable yet unavoidable, unless a concrete bound on the gradient's $L_2$ norm can be provided through alternative means. Tackling this challenge is a key contribution of our work. \newline     
\paragraph{Individual DP}
A recent line of work is closely related to the above-mentioned topic \cite{feldman2020individual, lecuyer2021practical}. Conceptually, these works attempt a more precise characterisation of individual privacy loss (termed \textit{individual differential privacy, IDP}) which is used to e.g. automatically halt the utilisation of the individual's data in an analysis while the data of other individuals can still be used (\textit{DP filtering}). Notably the work by \cite{feldman2020individual} also relates individual privacy guarantees to the \textit{Lipschitz} constant of the function which is applied to the data.

\section{System Description}
\label{hybrid_autodiff}
We begin by describing the main components of the proposed framework. Our system consists of a \textbf{front-end} user-facing component and of a tensor manipulation library similar to common machine learning and linear algebra libraries. The user specifies a series of computations to be performed (e.g. defining a neural network architecture) as well as the inputs to the computation (e.g. input tensors, weights, etc.). Two key principles apply at the first step of the computation: (1) The inputs can be specified \textit{abstractly}, that is, with a symbolic name (e.g. $\mathit{x_0}$) and (2) operations are per default defined in terms of \textit{individuals}, and not as \textit{batches} of inputs. At any point in the graph specification the user can request the partial derivatives of an operation with respect to some arbitrary input. \newline
The second component of our system is a \textbf{compiler tool-chain} consisting of a \textit{pre-processor} which emits an intermediate representation (IR) of the specified computations and a \textit{compiler} which converts the IR into low-level code. This can occur \textit{just-in-time}, which immediately returns a reference to a byte-code object representing the compiled expression to the user (at the cost of decreased performance) or \textit{ahead-of-time}, where a function is statically compiled into an optimised binary for execution on an arbitrary computational back-end (at the cost of longer compilation times). By default, these computational \textit{kernels} can be executed on batches of inputs, thus recovering the functionality of other frameworks. We note that, up to this point, no actual numeric input is required. The final computation takes place by substituting the abstract inputs specified at the beginning, with concrete values. In the case of \textit{just-in-time} compiled expressions, partial evaluation can also take place, where only a subset of inputs are specified numerically. \newline
The system, like other contemporary AD systems, is capable of computing differentials of arbitrary functions, provided a valid sub-differential exists. However, as such operations may contain discontinuities (e.g. piece-wise or step functions), the compiler maintains a reference to all possible execution paths and converts discontinuous expressions into conditional statements while optimising all computations beyond them for memory economy.\newline

% Pure symbolic systems suffer from the expression swell, meaning that after a number of operations, the resulting functions grow in size significantly. Automatic differentiation systems are very efficient in calculating the gradients through employing the reverse mode automatic differentiation, however, they cannot determine explicit representations, making it possible to only numerically evaluate the properties of functions. We design a system that follows a hybrid approach to differentiation, compiling the symbolic expression of each individual gradient and their norms to function calls that can be efficiently evaluated. We highlight that any expression with a valid subgradient can be evaluated in this manner. However, conditional expressions and piecewise functions increase memory consumption as all execution paths have to be retained, therefore significantly reducing the performance of the computation.. 
\section{Experiments}
\label{experiments}
\subsection{Application to the DP analysis of database queries}
Our system can be utilised to determine the \textit{local Lipschitz constant} of an arbitrary differentiable function for calculating the associated DP guarantees, which we showcase in the setting of a fictional database query under Rényi DP (RDP) \cite{Mironov_2017}.\newline
Assume an analyst wishes to calculate the (\textit{fictional}) quantity \textit{age-adjusted body mass index}, calculated from an individual's age $a$, weight $w$ and height $h$ as follows:
\begin{equation}
\textrm{B}(a, w, h) = \frac{a*w}{h^2}
\label{adjusted_bmi}
\end{equation}
It is obvious that (\ref{adjusted_bmi}) is not globally \textit{Lipschitz} continuous. It is however possible to determine a \textit{local Lipschitz constant} given bounds on the inputs, which can be shown to correspond to
\begin{equation}
    K = \sup \Vert\nabla\textrm{B}\Vert_2
\end{equation}
and which we can determine by automatic differentiation. We receive:
\begin{equation}
 K = \sup \left ( \frac{1}{h^2}\sqrt{w^2+a^2+\frac{4a^2w^2}{h^2}} \right) \bigg |_{(w, a, h)}
\end{equation}
Constrained minimisation of the inverse of this expression with reasonable bounds on the inputs given prior knowledge yields the desired quantity which can be used to compute the RDP guarantee as $(\alpha, \frac{\alpha K^2}{2\sigma^2})$. \newline
 
Of note, this method relies on retrieving a \textit{global} optimum for the expression, which can either be obtained in closed form by higher-order differentiation or algorithmically, e.g. through the use of \textit{simplicial homology global optimization}  \cite{endres2018simplicial}. Alternatively, methods for approximate \textit{Lipschitz}-constant estimation such as \cite{Scaman2018-gr} can be utilised.
 
\subsection{Application to DP-SGD}
In Algorithm \ref{alg:dpsgd}, we present a modified version of DP-SGD enabled by obtaining closed-form bounds on the maximum $L_2$-norm of the gradients, realised through the \textit{ahead-of-time} compilation of a closed-form-expression for the gradient norm using our framework. This allows the following distinct options for training with DP-SGD: 
\begin{enumerate}
    \item It is possible to compute the maximum sensitivity of the gradient given specific bounds on the function inputs (features, weights, biases etc.). This calculation can be performed ahead of the actual training training and utilised to guide network architecture design decisions. We note that, for complex functions, estimating this constant precisely may be very hard and some bound with an acceptable effect on privacy guarantees may have to be used instead \cite{fazlyab2019efficient}. Provided a reasonable bound exists, this method can be used to replace gradient clipping during training: Recall that the maximum sensitivity can be computed given bounds on all function inputs. Hence, the sensitivity will also remain bounded, as long as the function inputs satisfy the pre-specified constraints. During training, data and targets are typically normalised to a certain range. This renders the realised values of the weights at every iteration the sole determinant of sensitivity. By re-normalising the weights between to values, or clipping weights which exceed a pre-specified bound, it is thus possible to avoid gradient clipping altogether. However, bounding the weights may also introduce bias or hinder training by over-regularisation.
    \item For simple architectures, it is possible to determine the maximum sensitivity at every iteration. Samples whose norm remains below the maximum sensitivity require no clipping, and the noise can be scaled to the \textit{true} sensitivity value. This process is similar to the technique described by \cite{feldman2020individual} and \cite{lecuyer2021practical}. However, a costly optimisation procedure is necessary at every step, which does not scale appropriately. 
    \item To combine advantages of both approaches, it has been recommended to utilise activation functions which naturally bound the sensitivity, as described in \cite{papernot2020tempered}. Our framework allows to determine the exact \textit{Lipschitz} constant in advance (given potential additional bounds on the inputs or weights), without requiring step (2) to be executed at every iteration.
\end{enumerate}

We note that our system \textit{naturally} provides \textit{per-sample} gradients, as operations are specified per individual and are only dispatched at runtime. Thus, our framework allows to mitigate the performance degradation associated with obtaining per-sample gradients or their norms from batches of inputs.

\begin{algorithm}[h!]
  \caption{DP-SGD with hybrid automatic differentiation}
  \label{alg:dpsgd}
\begin{algorithmic}
  \STATE {\bfseries Input:} Examples $\{x_i,\ldots,x_n\}$, Loss function $\mathcal{L}(\theta) = \frac{1}{N}\sum_i\mathcal{L}(\theta, x_i)$. Parameters: learning rate $\eta_t$, noise scale $\sigma_t$, group size L, gradient norm bound C.
  \STATE Initialize $\theta$ randomly. \vspace{2pt}
  
  \STATE {\bfseries Pre-compute and compile gradient/ gradient norm:}
  \STATE Compute $\mathbf{g}_t(x_i, \theta)\leftarrow \nabla_{\theta_t}\mathcal{L}(\theta_t, x_i)$ symbolically
  \STATE Compute $\Vert\mathbf{g}_t(x_i, \theta)\Vert_2$ symbolically \vspace{2pt}
  \STATE {\bfseries Compute Lipschitz constant $K$:}\newline
  (1) Design architecture using suitable bounded activation functions.\newline
  (2) If a lower bound on $K$ is required, determine $K$ := $ \sup \Vert\mathbf{g}_t(x_i, \theta)\Vert_2$ under required constraints.
%   \STATE {\bfseries Define subroutines:}
%   \STATE $\textrm{OP CLIP}(g, C):= \mathbf{g}/\max(1, \frac{\Vert\mathbf{g}_t(x_i)\Vert_2}{C})$ \vspace{2pt}
%   \STATE $\textrm{OP FILT}(x_i, g_{max})$ := 
%   \IF{privacy budget exhausted}
%   \STATE exclude $x_i$ from further training
%   \ELSE
%   \STATE add privacy spent
%   \ENDIF \vspace{2pt}
  \FOR{$t \in [T]$}
  \STATE Take a random sample $L_t$ with sampling probability $\frac{L}{N}$ (e.g. \textit{Poisson} or uniform sampling)
  \FOR{each $i\in L_t$}
  
%   \IF{$g_{max} > C$}
%   \STATE $\textrm{OP CLIP}(g, C)$ \textbf{or} $\textrm{OP FILT}(x_i, g_{max})$ 
%   \ENDIF
  \STATE {\bfseries Substitute concrete values}
  \STATE Compute $\mathbf{g}_t(x_i, \theta)\leftarrow \nabla_{\theta_t}\mathcal{L}(\theta_t, x_i)$ given $x_i, \theta$
  \STATE {\bfseries Ensure bounded maximum sensitivity}
  \IF{lower bound on $K$ is required}
  \STATE enforce pre-computed bound on $K$ by bounding weights or gradient.
  \ENDIF
  \STATE {\bfseries Add noise}\newline
  $\widetilde{\mathbf{g}}\leftarrow \frac{1}{L}(\sum_i\mathbf{g}_t(x_i)+\mathcal{N}(0,\sigma^2K\mathbf{I})$
  \STATE {\bfseries Descent}\newline
  $\theta_{t+1}\leftarrow \theta_t - \eta_t\widetilde{\mathbf{g}}_t$
  \ENDFOR
  \ENDFOR
\end{algorithmic}
\end{algorithm}

\section{Discussion and conclusion}
\label{discussion}

Our work demonstrates that closed-form representations of arbitrary differentiable function compositions over private data can be utilised for the modelling of sensitivity in private machine learning. Our framework fits into a larger ecosystem of tools targeting automated sensitivity analysis \cite{traskkritika} and privacy budgeting through composition analysis of heterogenous mechanisms \cite{wang2019subsampled}. It moreover allows decoupling network/algorithm design from data analysis, a \textit{desideratum} in privacy-by-design workflows. We sketch the utilisation of our system in the context of IDP and DP-SGD, however view it as a useful component in the toolkit of adversarial machine learning researchers, whom it can assist in reasoning concretely over the mathematical properties of the model. Moreover, as our framework allows the derivation of analytic expressions for every input to the function, it can be used to quantify the impact of individual features on overall privacy loss. We intend to investigate the benefits of such feature-level analyses in future work. \newline
A notable limitation of the current form of our system is memory consumption and computational performance. Static compilation times scale quadratically with the number of parameters in the network. We note that, after compilation, execution times of the \textit{kernels} are constant for a given batch size. Currently, the compilation of a 2.5M parameter network (such as \textit{MobileNetV3}) would require approximately 60 hours on a single CPU core. Scaling our system to larger network architectures will require compiler-level optimisations such as multi-threaded compilation and employment of term simplification strategies such as \textit{common subexpression elimination}, which we outline as future work.\newline
In conclusion, hybrid automatic differentiation allows obtaining analytical expressions for quantities of interest in differentiable function compositions over private data such as neural networks. We are hopeful that our work will stimulate research at the intersection of compiler engineering, differentiable programming and privacy preserving machine learning to yield specialised tooling capable of efficiently manipulating such analytical representations to address key challenges in differentially private machine learning.

\bibliography{main.bib}
\bibliographystyle{icml2021}

% \appendix
% \section{Do \emph{not} have an appendix here}

% \textbf{\emph{Do not put content after the references.}}
% %
% Put anything that you might normally include after the references in a separate
% supplementary file.

% We recommend that you build supplementary material in a separate document.
% If you must create one PDF and cut it up, please be careful to use a tool that
% doesn't alter the margins, and that doesn't aggressively rewrite the PDF file.
% pdftk usually works fine. 

% \textbf{Please do not use Apple's preview to cut off supplementary material.} In
% previous years it has altered margins, and created headaches at the camera-ready
% stage. 
%%%%%%%%%%%%%%%%%%%%%%%%%%%%%%%%%%%%%%%%%%%%%%%%%%%%%%%%%%%%%%%%%%%%%%%%%%%%%%%
%%%%%%%%%%%%%%%%%%%%%%%%%%%%%%%%%%%%%%%%%%%%%%%%%%%%%%%%%%%%%%%%%%%%%%%%%%%%%%%

\end{document}